\documentclass[10pt,twocolumn,letterpaper]{article}

\usepackage{iccv}
\usepackage{times}
\usepackage{epsfig}
\usepackage{graphicx}
\usepackage{amsmath}
\usepackage{amssymb}
\usepackage{multirow}
\usepackage{algorithm}
\usepackage{algorithmic}
\usepackage{booktabs}
\usepackage[ruled, lined, linesnumbered, commentsnumbered, longend, algo2e]{algorithm2e}

\usepackage[breaklinks=true,bookmarks=false]{hyperref}

\iccvfinalcopy 

\def\iccvPaperID{7196} 

\ificcvfinal\fi

\begin{document}

\title{An Adaptive Model Ensemble Adversarial Attack for \\ Boosting Adversarial Transferability}

\author{Bin Chen$^1$ \qquad Jiali Yin$^1$ \qquad  Shukai Chen$^1$ \qquad  Bohao Chen$^2$ \qquad Ximeng Liu$^1$\\
$^1$Fuzhou University, Fujian, China \qquad  $^2$Yuan Ze University, Taipei, Taiwan\\
{\tt\small c\_chenbin@foxmail.com,~jlyin@fzu.edu.cn,~chenshukai770@163.com,~\{hd840207, snbnix\}@gmail.com}\\
}

\maketitle
\ificcvfinal\thispagestyle{empty}\fi

\begin{abstract}
While the transferability property of adversarial examples allows the adversary to perform black-box attacks (\textit{i.e.}, the attacker has no knowledge about the target model), the transfer-based adversarial attacks have gained great attention. Previous works mostly study gradient variation or image transformations to amplify the distortion on critical parts of inputs. These methods can work on transferring across models with limited differences, \textit{i.e.}, from CNNs to CNNs, but always fail in transferring across models with wide differences, such as from CNNs to ViTs. Alternatively, model ensemble adversarial attacks are proposed to fuse outputs from surrogate models with diverse architectures to get an ensemble loss, making the generated adversarial example more likely to transfer to other models as it can fool multiple models concurrently. However, existing ensemble attacks simply fuse the outputs of the surrogate models evenly, thus are not efficacious to capture and amplify the intrinsic transfer information of adversarial examples. In this paper, we propose an adaptive ensemble attack, dubbed AdaEA, to adaptively control the fusion of the outputs from each model, via monitoring the discrepancy ratio of their contributions towards the adversarial objective. Furthermore, an extra disparity-reduced filter is introduced to further synchronize the update direction. As a result, we achieve considerable improvement over the existing ensemble attacks on various datasets, and the proposed AdaEA can also boost existing transfer-based attacks, which further demonstrates its efficacy and versatility. 
\end{abstract}

\section{Introduction}\label{section0}

\begin{figure}[t]
\begin{center}
\includegraphics[width=1.0\linewidth]{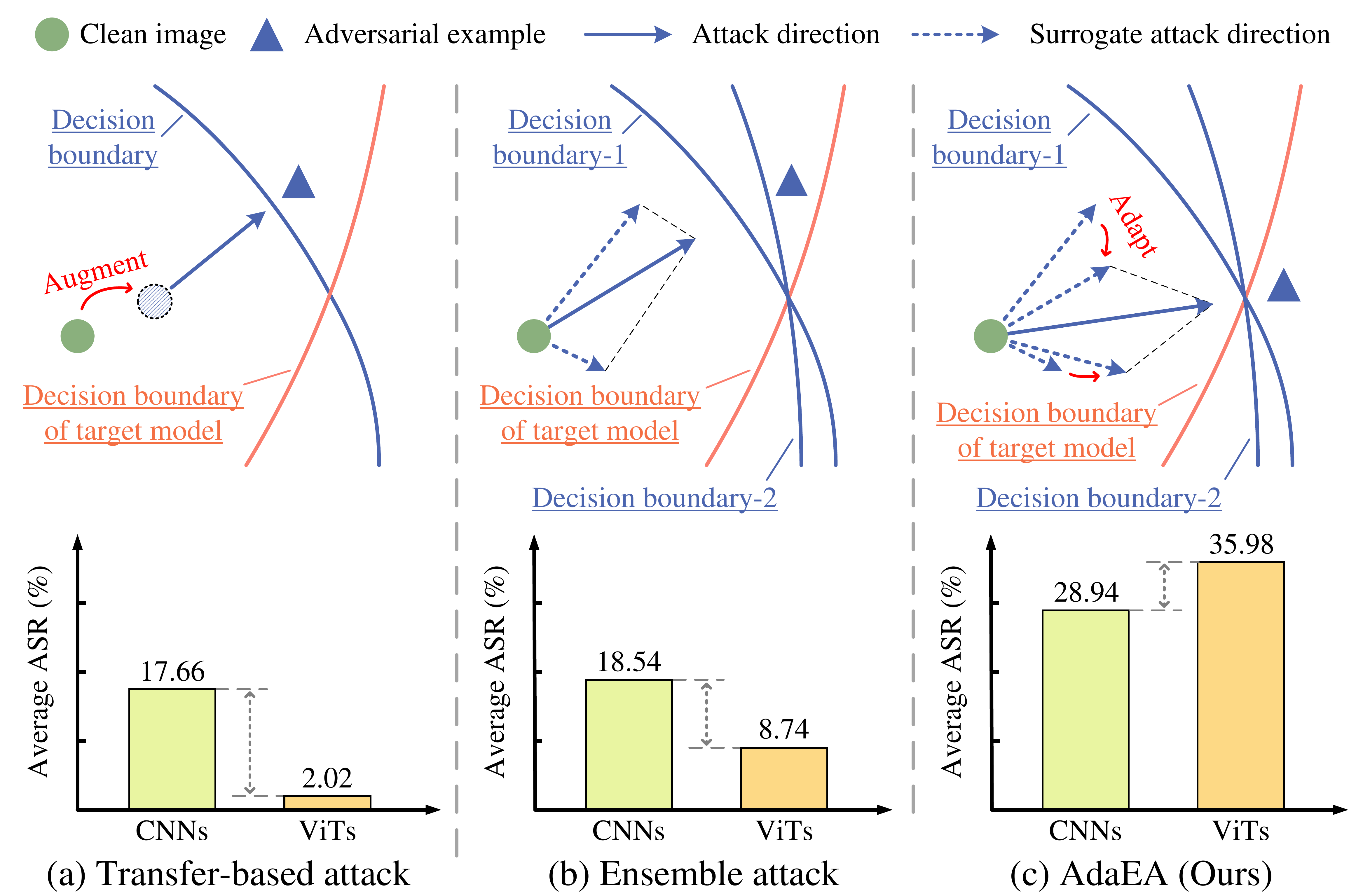}
\end{center}
\vspace{-5pt}
\caption{Overview of different attack schemes and performance. (a) Transfer-based methods strengthen the critical parts in images to improve the attack transferability, but fail to transfer across DNNs with wide differences due to the limited adversarial information. (b) Model ensemble attacks integrate multiple surrogate models for finding the more transferable attack, but existing works generally neglect the individual characteristics of each model, leading to under-optimal results. (c) Our AdaEA performs adaptive ensemble by amplifying the transferable information in each surrogate model and achieves remarkable improvements.}
\label{fig:introduction}
\vspace{-10pt}
\end{figure}

Deep neural networks (DNNs), including convolutional neural networks (CNNs)~\cite{ResNet_2016_CVPR, wrn_zagoruyko2016wide, bit_kolesnikov2020big} and vision transformers (ViTs)~\cite{ViT_16x16_2020image,DeiT_touvron2021training,swin_liu2021swin}, have brought impressive advances to the state-of-the-art across various machine-learning tasks. At the moment, however, they are found to be vulnerable to adversarial examples~\cite{AdvExample_szegedy2013intriguing}, \textit{i.e.}, adding imperceptible hand-crafted perturbations to the original inputs can lead to wrong prediction behavior of DNNs. This discovery arises severe security hazards in the deployment of DNNs. More importantly, some well-designed adversarial examples can transfer across models. That is, an adversarial example crafted from a surrogate model can also disturb other models. This property of adversarial examples, known as \textit{tranferability}, allows the adversary to attack a target model without knowing its interior, thus poses a more realistic threat to \textit{black-box} applications (\textit{i.e.}, the architectures and parameters are inaccessible to users). 

To set up the first step for improving model robustness and prevent potential threats from black-box attacks, the research on improving the transferability of adversarial examples has attracted wide attention in recent years. The attack transfer success rates vary depending on the difference between the surrogate and target models, the more similar the surrogate and target models are, the higher transfer success rate can be achieved. Thus a bunch of works have been proposed to improve the transferability of adversarial examples by maximizing the perturbation on critical parts that are shared among DNNs. The mainstream strategies include maximizing information from important neurons~\cite{Zhang_Neuron2022_CVPR, Wang_neuro_2021}, increasing input diversity~\cite{DI2-FGSM_xie2019improving,Byun_2022_CVPR}, and incorporating momentum~\cite{MIM_Dong_2018_CVPR,Tansfer_Mom_wang2021} into iterative-based attack. Despite their effectiveness, these methods always fail in transferring across models with wide architecture differences (\textit{i.e.}, CNNs and ViTs), as shown in Figure\,\ref{fig:introduction} (a).

Similar to traditional ensemble methods which draw on the wisdom of multiple weak learners with diverse predictions to improve the overall accuracy, a line of research proposes to utilize an ensemble of surrogate models to generate adversarial examples that can successfully attack all the surrogate models. Intuitively, the approach can improve the transferability of adversarial examples as it can potentially capture intrinsic transferable adversarial information since the adversary can fool several models with wide differences concurrently. Moreover, such an ensemble could also be easily incorporated with existing transfer-based adversarial attacks without confliction. Several model ensemble based methods have been explored~\cite{liu2017delving,BAST_ensemble_he2019new}, however, most of them only equally fuse the outputs (\textit{i.e.}, logits or losses) of all models to get an ensemble loss for applying gradient-based attack, which may limit the potential capability of the model ensemble attacks, as shown in Figure\,\ref{fig:introduction} (b). Although a recent work~\cite{Xiong_2022_CVPR} noticed the gradient variances among the surrogate models, the ensemble is still under-optimal due to the ignorance of individual characteristics of each model.

In this paper, we focus on the model ensemble adversarial attack for improving the transferability of adversarial examples. We first observe that simply averaging the outputs of ensemble models ignore the advantages of each model, where the transferable information captured from one model can be smoothed by another model during the fusion process, thus leading to the under-optimized results. To cope with this problem, we propose to adptively ensemble the outputs of each model via the {adaptive gradient modulation} (AGM) strategy. Specifically, we define the \textit{adversarial ratio} to evaluate the contribution discrepancy among the surrogate models to the overall adversarial objective, which is then exploited to adaptively modulate the gradient fusion, offering more efforts on the amplification of transferable information in the generated adversarial examples. Moreover, the ensemble gradient may greatly differ or even oppose with the individual gradient of surrogate models, which has been proven to have a correlation with the overfitting problem in ensemble~\cite{Xiong_2022_CVPR}. Hence, we further introduce a {disparity-reduced filter} (DRF) where a disparity map is computed to reduce the variances among surrogate models and synchronize the update direction. Finally, the adversarial transferability could be enhanced by applying the above two mechanisms, as demonstrated in Figure\,\ref{fig:introduction}. We term the proposed method as adaptive esnemble attack (AdaEA), and perform extensive experiments on diverse datasets to validate that our AdaEA can consistently outperform the existing methods. To sum up, the key contributions of this work are three-fold:
\begin{itemize}
    \item We propose an adaptive ensemble adversarial attack, dubbed AdaEA, which offers a more comprehensive ensemble attack for a broad class of models with wide architecture differences, such as CNNs and ViTs.
    \item Our AdaEA views the ensemble attack from the gradient optimization perspective, and controls the optimization process via AGM strategy as well as reducing the disparity by DRF to synchronize the optimization direction.
    \item The proposed AdaEA can not only largely enhance the ensemble effectiveness compared to existing ensemble methods, but also consistently improve the attack performance when incorporated with the existing transfer-based gradient attacks.
\end{itemize}

\section{Related Works}
\subsection{Adversarial Attacks}
Since Szegedy \textit{et~al.}~\cite{AdvExample_szegedy2013intriguing} first reported the existence of adversarial examples, extensive efforts have been devoted to highlighting the vulnerability of DNNs. An adversarial attack usually produces adversarial examples by adding a perturbation $\delta$ to an original input image $x$ with the objective that can make the model discriminative loss $\mathcal{L}$ maximized, \textit{i.e.}, $\arg\max_{x+\delta}\mathcal{L}(f(x+\delta), y)$. To make the perturbation imperceptible, the perturbation $\delta$ is subject to a constraint $\mathcal{S}$, which is defined as $\mathcal{S}=\{ \|\delta\|_{p}\leq \epsilon\}$ by the given $\ell_p$-norm distance and the maximum strength $\epsilon$.

\vspace{3pt}
\noindent\textbf{Gradient-based adversarial attacks.} To optimize the attack objective, the gradient information are usually used to  maximize the model loss. Goodfellow~\textit{et~al}\mbox{.}~\cite{FGSM_goodfellow2014explaining} designed a Fast Gradient Sign Method (FGSM) to produce strong adversarial examples based on the investigation of CNN linear nature. Wang~\textit{et~al}\mbox{.}~\cite{BIM_kurakin2016adversarial} and Madry~\textit{et~al}\mbox{.}~\cite{pgd_madry2017towards} further broke the one-step generation of perturbation in FGSM into iterative generation and proposed I-FGSM and Projected Gradient Descent (PGD) attack. While these attacks can exhibit high attack success rate on the white-box models, they usually reveal low transfer rate to black-box models since the gradients information is hard to approximate.

\noindent\textbf{Transfer-based adversarial attacks.}  To improve the transferability, existing works try to maximize the distortion on the critical parts of inputs. Wang~\textit{et~al}\mbox{.}~\cite{Wang_neuro_2021} and Zhang~\textit{et~al}\mbox{.}~\cite{Zhang_Neuron2022_CVPR} investigated the distortion on features based on the importance of neural in the DNNs. Xie~\textit{et~al}\mbox{.}~\cite{DI2-FGSM_xie2019improving} and Dong~\textit{et~al}\mbox{.}~\cite{Dong_2019_CVPR} incorporated the FGSM with either input diversity or translation-invariant strategies to produce diverse input patterns for generation of adversarial examples. Gao~\textit{et~al}\mbox{.}~\cite{PI-FGSM_gao2020patch} proposed the PI-FGSM which generates patch-wise perturbation rather than pixel-wise, that is beneficial for black-box attack. Although these attacks can achieve transferability improvements over the primordial gradient-based attacks, they can hardly transfer to the new architecture of DNNs, \textit{i.e.}, the ViT family. 

\begin{figure}[t]
\begin{center}
\includegraphics[width=0.90\linewidth]{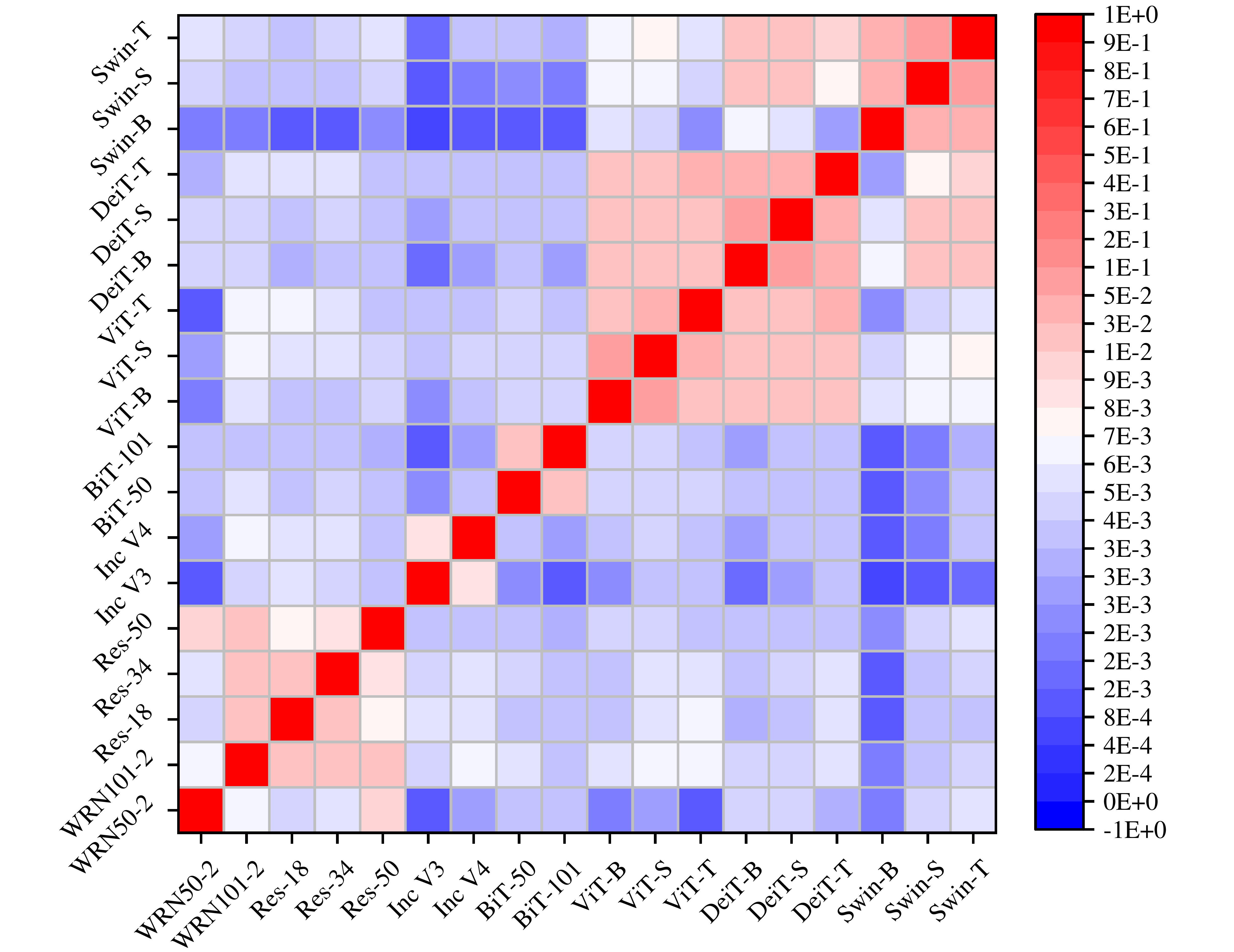}
\end{center}
\vspace{-10pt}
\caption{Visualization of the cosine similarity between the gradients produced from different models. Note that the gradients are closer when the model architectures are more similar. }
\vspace{-15pt}
\label{fig:grad_cam}
\end{figure}

\noindent\textbf{Model ensemble attacks.} Ensemble attack methods usually craft adversarial examples by performing a weighted linear sum of the multiple white-box attacks in parallel. Liu~\textit{et~al}\mbox{.}~\cite{liu2017delving} directly averaged the predictions of multiple modes to get an ensemble loss for applying gradient-based attack.
Dong~\textit{et~al}\mbox{.}~\cite{Dong_2019_CVPR} further fused the logits and losses of ensemble models.
Xiong~\textit{et~al}\mbox{.}~\cite{Xiong_2022_CVPR} noticed the variance among the ensemble models and proposed a stochastic variance-reduced ensemble (SVRE) attack to improve the attack generalization. While improvements being achieved, the ensemble is still under-optimal due to the less investigation in the individual advantages of each model.

\subsection{Adversarial Defenses}
As the counterpart of adversarial attack, enormous efforts have been proposed to defend against adversarial examples, which generally fall into two categories. 
The first is referred to as adversarial training~\cite{pgd_madry2017towards,Cai2018Curriculum,Zhang2020ES,ensemble_attack_and_defence_tramr2018ensemble,pang2019improving,kariyappa2019improving,Yang_DVERGE}, which is regarded as the most reliable and effective method. Its key idea is to leverage the online generated adversarial examples into the training dataset so that the model can prefer more robust features during learning~\cite{pgd_madry2017towards}. To improve the defense efficiency, state-of-the-art methods further propose to incorporate curriculum attack generation~\cite{Cai2018Curriculum}, early stopping~\cite{Zhang2020ES}, and ensemble schemes~\cite{ensemble_attack_and_defence_tramr2018ensemble,pang2019improving,kariyappa2019improving,Yang_DVERGE}.
The second line of adversarial defense is input transformation-based methods, which aim to eliminate the adversarial information from adversarial examples by preprocessing. Many state-of-the-art defense methods for defending against adversarial examples have been proposed, including denoising images with high-level representation~\cite{Liao2018denoise}, randomly resizing~\cite{r&p} and smoothing~\cite{RS}, compressing input image~\cite{ComDefend, jpeg, bit-r, FD} and purifying the input images using neural network~\cite{NRP}.
In this paper, we employ these state-of-the-art defenses to evaluate the effectiveness of our attack method.

\begin{figure*}[t]
\begin{center}
\includegraphics[width=1.0\linewidth]{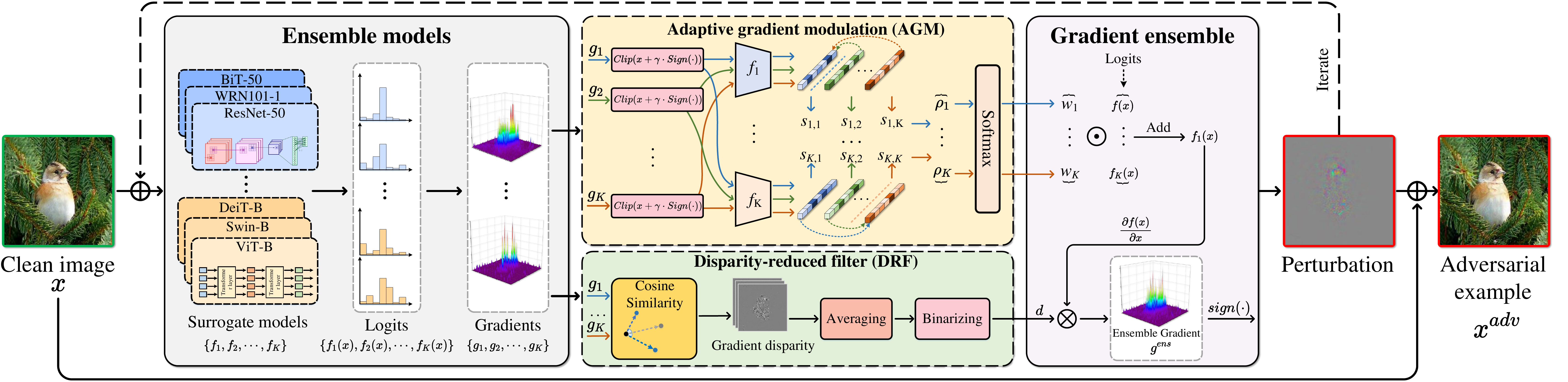}
\end{center}
\caption{An overview of our AdaEA. The gradients obtained from CNNs and ViTs are feed into the AGM and DRF to get the ensemble gradient for generating adversarial examples with gradient-based attack. }
\label{fig:framework}
\vspace{-5pt}
\end{figure*}

\section{Methodology}
\subsection{General Overview}
Improving the transferability of adversarial examples aims to make an adversarial example generated from a white-box surrogate model stay adversarial to hold-out black-box models. Typically, using a gradient-based method to iteratively find the optimal perturbation for a white-box model can be given by:
\begin{align}\label{eq:at}
	{x}^{adv}_{t+1} & = {x}^{adv}_{t}+\alpha\operatorname{sign}(\nabla_{{x}^{adv}_{t}} \mathcal{L}(f({x}^{adv}_{t}), y)),
\end{align}
where $\operatorname{sign}(\cdot)$ is the sign function, $\alpha$ is the step size, and $\nabla_{{x}^{adv}_{t}}\mathcal{L}$ denotes the gradient of the loss function $\mathcal{L}$ w.r.t. ${x}^{adv}_{t}$. Note that ${x}^{adv}_{1}$ is set to be $x$, and the final adversarial example is obtained by ${x}^{adv}_{T}$, $T$ is the iteration number. Intuitively, it can achieve high attack successful rate under the white-box setting, where $\nabla_{{x}^{adv}_{t}}\mathcal{L}$ is known. However, when transferred to a black-box model in which the $\nabla_{{x}^{adv}_{t}}\mathcal{L}$ is unknown, the attack successful rate would be dropped since the gradients are diverse in different models, as shown in Figure\,\ref{fig:grad_cam}. In particular, when the model architectures significantly differ, such as ViTs and CNNs, the gradients are extremely different, leading to a lower transfer attack rate.

To make the generated adversarial examples adversarial to a broad class of models, the ensemble attack is an effective strategy to enhance the attack transferability. The basic idea is to utilize the outputs of multiple white-box models to obtain the averaged model loss, and then the gradient-based attack is applied to generate the adversarial example. It transforms Eq.\,\eqref{eq:at} into the following representation:
\begin{align}\label{eq:ens_at}
	{x}^{adv}_{t+1} & = {x}^{adv}_{t}+\alpha\operatorname{sign}(\nabla_{{x}^{adv}_{t}} \mathcal{L}(\sum \limits_{k=1}^K w_k f_k({x}^{adv}_{t}), y)),
\end{align}
where $w_k$ is the ensemble weights for $k$-th surrogate model $f_k$, $\forall w_k\ge0$ and $\sum\nolimits_{k=1}^{K}{{{w}_{k}}}= 1$; and $K$ is the number of surrogate models. 
Existing ensemble methods generally average the logits~\cite{liu2017delving}, predicted probabilities~\cite{Dong_2019_CVPR}, or losses~\cite{Dong_2019_CVPR} of surrogate models to obtain the ensemble loss for generating gradient information. However, such simple ensemble ignores the individual variance across the surrogate models, thus significantly limits the overall attack performance. Let us take Figure\,\ref{fig:grad_cam} as an example again, as the gradients vary across different models, directly equally merging the outputs of models would lead to under-optimal results since the adversarial information captured by each model is not evaluated and amplified. 

\subsection{Adaptive Ensemble Adversarial Attack}
In this work, we focus on the model ensemble methods following Eq.\,\eqref{eq:ens_at}. 
Instead of directly averaging the outputs of surrogate models as the previous works, we propose AdaEA equipped with AGM and DRF mechanisms to amend the gradient optimization process for boosting the transferable information in the generated adversarial examples. 
Specifically, AGM first modulates the gradient of each ensemble model by the defined \textit{adversarial ratio} which identifies the contribution discrepancy of each surrogate model to the overall adversarial object, and then the DRF further synchronizes the gradient update direction by filtering out the disparity part of ensemble gradients. An overview of AdaEA is shown in Figure\,\ref{fig:framework}.

\vspace{5pt}
\noindent\textbf{Adaptive gradient modulation.} After obtaining the outputs $f_i(x)$ and gradient information $g_i$ from each surrogate model $f_i$ by feeding the input image, \textit{i.e.}, $g_i=\nabla_{x^{adv}_t} \mathcal{L}(f_i(x^{adv}_t), y)$, we propose to adaptively modulate the model ensemble via monitoring the discrepancy of their contributions to the adversarial attack objective. Specifically, for the $i$-th ensemble model $f_i$, we evaluate the potential adversarial transferability in the $g_i$ by testing the attack performance of adversarial examples generated from $g_i$ on other models, which we define as \textit{adversarial ratio}, and then adjust the ensemble weight based on the adversarial ratio of each model. Here we first conduct the testing process by computing:
\begin{equation}
    s_{k,i}= - \textbf{1}_y \cdot \log \left(\text{softmax}\left(\textbf{p}_k[x_{t}^{adv}+\alpha\operatorname{sign}(g_i)]\right)\right)~,
    \label{eq:weight_balance_1}
\end{equation}
where $\textbf{p}_k (\cdot)$ denotes the logits output from $f_k$, and $\textbf{1}_y$ is the ground truth logits. $s_{k,i}$ can be considered as the $k$-th model loss on the adversarial example generated by using the gradient from $i$-th model. We then define the adversarial ratio $\rho_i$ as:
\begin{equation}
    \rho_i= \frac{\beta}{K-1} \sum \nolimits_{k=1, k\neq i}^K \frac{s_{k,i}}{s_{k,k}}~,
    \label{eq:weight_balance_2}
\end{equation}
where $\beta$ denotes a hyperparameter that controls the effect of ensemble weighting, which is further discussed in Sec.\,\ref{sec:ablation}. Note that a higher value of $\rho_i$ denotes a better transfer attack of adversarial example generated from $g_i$, implying that $g_i$ contains more transferable adversarial information. By doing so, we can figure out which model can provide more generic adversarial information and adaptively assign a higher ensemble weight. Thus, according to the adversarial ratio of each model, we use a softmax function to normalize the ensemble weight of each model by:
\begin{equation}
    w_1^*, w_2^*,..., w_K^*= \text{softmax}(\rho_1, \rho_2,..., \rho_K).
    \label{eq:weight_balance_3}
\end{equation}

With the obtained $w_i^*$, the output of each surrogate model with more potential adversarial transferability information is amplified in the ensemble gradient of Eq.\,\eqref{eq:ens_at}, thus leading to a higher transfer attack success rate on the hold-out black-box models.

\vspace{5pt}
\noindent\textbf{Disparity-reduced filter.} As discussed, the gradient optimization direction of surrogate models vary tremendously in a big range, sometimes the gradients walk towards direction against each other and the result leads to an overfit to the ensemble model~\cite{Xiong_2022_CVPR}. To solve the problem and synchronize the update direction, we introduce an extra disparity-reduced filter to reduce the gradient variances among surrogate models. We first apply the cosine similarity to evaluate the deviation of gradients in surrogate models, and compute the disparity map $d_i$ by averaging the similarity score with gradients of other models, which can be described as follows:
\begin{align}\label{eq:drf_1}
    d_i^{(p,q)} &= \frac{1}{K - 1}  \sum \nolimits_{k=1, k \neq i}^K \text{cos}\left(\overrightarrow{g}_i^{(p,q)}, \overrightarrow{g}_k^{(p,q)} \right) ~, 
\end{align}
where $\text{cos}(\cdot)$ denotes cosine similarity function, $\overrightarrow{g}_i^{(p,q)}$ and $\overrightarrow{g}_k^{(p,q)}$ denote the vector extracted from the position $(p,q)$ through channels of gradient $g_i$ and $g_k$, respectively. The final disparity map $d$ for ensemble gradients is obtained by averaging all the $d_i$. We then clean the disparity part in the ensemble gradient by using a filter $\mathbf{B}$ as:
\begin{align}
\label{eq:drf_3}
\mathbf{B}{(p,q)} = 
\begin{cases}
0, ~~~\text{if }d_i^{(p,q)} \leq \eta \\
1,  ~~~\text{otherwise}
\end{cases},
\end{align}
where $\eta$ is the tolerance threshold for the disparity filtering. By filtering out the disparity part of the ensemble gradients, the gradient optimization direction can be synchronized. To this end, the ensemble gradient can be obtained by rewriting Eq.\,\eqref{eq:ens_at} as:
\begin{align} 
    \label{eq:ensemble_gradient_1}
   &g_{t+1} = \nabla_{x^{adv}_t} \mathcal{L}(\sum\nolimits_{k=1}^{K} w_k^* f_k(x^{adv}_t), y)\otimes \mathbf{B},
\end{align}
where $\otimes$ denotes the element-wise multiplication. Hence, the disparity among the surrogate models can be suppressed. We provide more discussions about DRF in terms of both qualitative and quantitative analysis in the supplementary material. The overall AdaEA procedure is shown in Algorithm\,\ref{alg:AdaEA}.

\section{Experiments}
\subsection{Experimental Setting}\label{sec:ex_settings}
\noindent \textbf{Datasets.} We conduct experiments on CIFAR-10, CIFAR-100 and ImageNet datasets \cite{cifar10,imagenet} which are widely used in both classification and adversarial attack tasks~\cite{Xiong_2022_CVPR, liu2017delving}.

\noindent \textbf{Networks.} We choose target models from both branches of CNNs and ViTs for the black-box attack task, including ResNet-50 (Res-50)~\cite{ResNet_2016_CVPR}, WideResNet-50 (WRN-50)~\cite{wrn_zagoruyko2016wide}, BiT-M-R50$\times$1 (BiT-50)~\cite{bit_kolesnikov2020big} and BiT-M-R101 (BiT-101)~\cite{bit_kolesnikov2020big} in CNN branch; and ViT-Base (ViT-B)~\cite{ViT_16x16_2020image}, DeiT-Base (DeiT-B)~\cite{DeiT_touvron2021training}, Swin-Base (Swin-B)~\cite{swin_liu2021swin} and Swin-Small (Swin-S)~\cite{swin_liu2021swin} in ViT branch. 
As for surrogate models, we choose ResNet-18 (Res-18)~\cite{ResNet_2016_CVPR}, Inception v3 (Inc-v3)~\cite{inception_szegedy2014going}, ViT-Tiny (ViT-T)~\cite{ViT_16x16_2020image} and DeiT-Tiny (DeiT-T)~\cite{DeiT_touvron2021training} in the later experiments by default.

\begin{algorithm}[!t]
    \caption{The AdaEA algorithm}
    \label{alg:AdaEA}
    \KwIn{Input $(x,y)$, a list of $K$ surrogate models. Maximum range of perturbation $\epsilon$, the step size of iteration attack $\alpha$, and the number of iterations in the inner gradient-based attack $T$.}
    \BlankLine
    
    \KwOut{Adversarial example $x^{adv}$.}
    $x^{adv}_1 \leftarrow x$ \par
    \For{$t \leftarrow 1$ \KwTo $T$}{
        \# Calculating the gradients of all $K$ models \par
        $g_k \leftarrow \nabla_{x^{adv}_t} \mathcal{L}(f_k(x^{adv}_t), y)$ \par

        \# Performing adaptive gradient modulation \par
        Compute the adversarial ratio $\rho_i$ of each model using Eqs.\,\eqref{eq:weight_balance_1}-\eqref{eq:weight_balance_2} \par
        Compute the weight $w$ for each model using Eq.\,\eqref{eq:weight_balance_3} \par
        \# Performing disparity-reduced filter \par
        Compute the disparity map $d$ using Eq.\,\eqref{eq:drf_1} \par
        \# Ensemble the gradient \par 
        Compute the gradient $g_{t+1}^{ens}$ using Eqs.\,\eqref{eq:drf_3}-\eqref{eq:ensemble_gradient_1}.\par 
        \# Updating adversarial example \par 
        $x^{adv}_{t+1} \leftarrow \text{Clip}_x^\epsilon \{ x^{adv}_t + \alpha\operatorname{sign}(g^{ens}_{t+1}) \}$
    }
    $x^{adv} \leftarrow x^{adv}_T$
\end{algorithm}

\noindent \textbf{Comapred methods.} Two pioneering ensemble attack methods, \textit{i.e.}, Ens~\cite{liu2017delving} and SVRE~\cite{Xiong_2022_CVPR}, are employed as baselines to compare with our AdaEA. All the ensemble methods follow the same ensemble settings in experiments.

\begin{table*}[t]
\centering
\fontsize{8.5}{10.5}\selectfont
\caption{The black-box attack success rate (\%) against eight naturally trained models. The bolded numbers indicate the best results and $\Delta$ represents the improvements over the baseline.}
\vspace{5pt}
\label{tab:comparison}
\begin{tabular}{ccccccccccc}
\toprule[1pt]
\multicolumn{1}{c}{Dataset} & \multicolumn{1}{c}{Attack} & Res-50 & WRN101-2 & BiT-50 & BiT-101 & ViT-B & DeiT-B & Swin-B & Swin-S & Average ($\Delta$) \\ 
\toprule[1pt]
\multirow{3}{*}{CIFAR-10}
 & Ens & 50.42 & 26.85 & 21.83 & 17.61 & 11.59 & 26.15 & 22.61 & 35.42 & 26.56 \\
 & SVRE & 54.08 & 28.47 & 23.28 & 19.06 & 13.83 & 31.00 & 25.17 & 40.17 & 29.38 (+2.82)  \\
 & \textbf{AdaEA} & \textbf{61.54} & \textbf{38.07} & \textbf{33.36} & \textbf{28.99} & \textbf{31.77} & \textbf{59.72} & \textbf{45.90} & \textbf{61.38} & \textbf{45.09 (+18.53)} \\ \midrule
\multirow{3}{*}{CIFAR-100} 
 & Ens & 80.13 & 67.89 & 60.79 & 44.78 & 45.46 & 69.50 & 64.40 & 77.14 & 63.76 \\
 & SVRE & 82.06 & 68.68 & 62.59 & 46.30 & 48.11 & 73.63 & 67.94 & 80.49 & 66.23 (+2.47) \\
 &\textbf{AdaEA} & \textbf{82.19} & \textbf{70.02} & \textbf{65.28} & \textbf{48.63} & \textbf{60.20} & \textbf{82.83} & \textbf{75.21} & \textbf{84.41} & \textbf{71.10 (+7.34)} \\ \midrule
\multirow{3}{*}{ImageNet} 
 & Ens & 52.90 & 58.10 & 56.86 & 48.27 & 39.94 & 51.38 & 25.95 & 37.66 & 46.38 \\
 & SVRE & \textbf{53.10} & 57.84 & 56.90 & 48.38 & 40.03 & 52.06 & 25.54 & 37.26 & 46.39 (+0.01)  \\
 &\textbf{AdaEA} & \textbf{53.10} & \textbf{58.33} & \textbf{58.57} & \textbf{50.06} & \textbf{46.13} & \textbf{58.05} & \textbf{29.37} & \textbf{41.30} & \textbf{49.36 (+2.98)} \\ 
 \bottomrule[1pt]
\end{tabular}
\end{table*}

\begin{table*}[ht]
\centering
\fontsize{8.5}{10.5}\selectfont
\caption{The attack success rate (\%) of adversarial examples generated by ensemble attacks based on different attack methods on CIFAR-10. The bolded numbers indicate the best results and $\Delta$ represents the improvements over the baseline.}
\vspace{5pt}
\label{tab:base_comparison}
\begin{tabular}{ccccccccccc}
\toprule[1pt]
Base & Attack & Res-50 & WRN101-2 & BiT-50 & BiT-101 & ViT-B & DeiT-B & Swin-B & Swin-T & Average ($\Delta$) \\ 
\toprule[1pt]
\multirow{3}{*}{FGSM~\cite{FGSM_goodfellow2014explaining}} 
 & Ens & 21.32 & 16.22 & 12.58 & 10.69 & 7.17 & 10.68 & 8.30 & 15.23 & 12.77 \\
 & SVRE & 26.05 & 20.61 & 21.26 & 18.87 & 17.84 & 22.23 & 17.66 & 25.99 & 21.31 (+8.54) \\
 &\textbf{AdaEA} & \textbf{32.96} & \textbf{31.41} & \textbf{34.35} & \textbf{32.57} & \textbf{38.40} & \textbf{45.83} & \textbf{35.82} & \textbf{43.78} & \textbf{36.89 (+24.12)} \\ \midrule
\multirow{3}{*}{I-FGSM~\cite{BIM_kurakin2016adversarial}} 
 & Ens & 50.42 & 26.85 & 21.83 & 17.61 & 11.59 & 26.15 & 22.61 & 46.93 & 28.00 \\
 & SVRE & 51.92 & 27.50 & 22.90 & 18.29 & 13.30 & 30.74 & 24.84 & 51.01 & 30.06 (+2.06)  \\
 &\textbf{AdaEA} & \textbf{61.54} & \textbf{38.07} & \textbf{33.36} & \textbf{28.99} & \textbf{31.77} & \textbf{59.72} & \textbf{45.90} & \textbf{70.77} & \textbf{46.27 (+18.27)} \\ \midrule
\multirow{3}{*}{MI-FGSM~\cite{MIM_Dong_2018_CVPR}} 
 & Ens & 55.10 & 33.89 & 29.68 & 25.28 & 20.96 & 42.12 & 31.30 & 58.20 & 37.07 \\
 & SVRE & 31.46 & 21.37 & 18.53 & 16.21 & 15.53 & 26.86 & 20.70 & 33.69 & 23.04 (-14.03) \\
 & \textbf{AdaEA} & \textbf{66.58} & \textbf{44.45} & \textbf{41.90} & \textbf{37.23} & \textbf{45.96} & \textbf{70.78} & \textbf{53.61} & \textbf{78.00} & \textbf{54.81 (+17.74)} \\ \midrule
\multirow{3}{*}{DI$^2$-FGSM~\cite{DI2-FGSM_xie2019improving}} 
 & Ens & 90.28 & 67.34 & 63.06 & 57.65 & 51.19 & 82.44 & 76.31 & 91.26 & 72.44 \\
 & SVRE & 39.30 & 32.12 & 29.78 & 27.41 & 26.82 & 36.99 & 35.35 & 40.20 & 33.49 (-38.95) \\
 & \textbf{AdaEA} & \textbf{91.49} & \textbf{74.08} & \textbf{72.26} & \textbf{68.83} & \textbf{66.96} & \textbf{89.23} & \textbf{84.48} & \textbf{95.20} & \textbf{80.32 (+7.88)} \\ 
\bottomrule[1pt]
\end{tabular}
\vspace{-4pt}
\end{table*}

\noindent \textbf{Implementation details.} 
For the baselines and our AdaEA, we use the I-FGSM with $20$ iterations under $l_\infty$ constraint as the basic attack method, and set $\epsilon = 8/255$ and $\alpha = 2/255$ during the adversarial example generation. As for hyperparameter, we set $\eta=-0.3$ in DRF and $\beta=10$ in AGM. The inner update time in SVRE is set to be $4$ following its default setting.  All the experiments were implemented using Pytorch on an Intel Xeon Sliver and a NVIDIA A6000 GPU with 48GB graph memory.

\subsection{Main Results}\label{sec:results}
\noindent\textbf{General attack performance.} We first compare the general attack performance of AdaEA with existing ensemble methods on the naturally trained models under the black-box setting on CIFAR-10/100 and ImageNet. 
Table\,\ref{tab:comparison} reports the attack results on a broad class of black-box models, including both CNNs and ViTs. As we can see, SVRE can slightly improve the attack performance by reducing gradient variance across models compared to the baseline Ens. The improvements in terms of average success rates are around $3\%$ on CIFAR datasets. In contrast, our AdaEA can improve the attack transfer rate by a large margin, where we achieve more than $15 \%$ averaging improvements over SVRE on CIFAR-10, demonstrating the effectiveness of our AdaEA in finding and amplifying the intrinsic adversarial information of inputs via the AGM-DRF strategies.

\noindent\textbf{Combinations with transfer-based attacks.} We then attempt to test the integration of the existing transfer-based attacks in our AdaEA. We additionally use FGSM, MI-FGSM~\cite{MIM_Dong_2018_CVPR}, and DI$^2$-FGSM~\cite{DI2-FGSM_xie2019improving} as the base attacks for ensemble, and summarize the results in Table\,\ref{tab:base_comparison}. The results show that the attack success rate significantly improves combined with our AdaEA regardless of base attacks. Specifically, for FGSM and I-FGSM, using our AdaEA improves the average transfer success rate around $20\%$. For MI-FGSM and DI$^2$-FGSM attacks, our method also achieves consistently improvements over the existing ensemble attacks by a large margin, 
further indicating the promising versatility of our proposed AdaEA.

\noindent\textbf{Attack advanced defense models.} We also evaluate AdaEA on attacking models with various advanced defenses, including adversarial training defenses and input transformation-based defenses. The results are summarized in Table~\ref{tab:comparison_defence}. For adversarial training defense, we use adversarial trained Inc-v3$_{ens3}$, Inc-v3$_{ens4}$ and Inc-v2$_{ens}$ networks as the target model following previous works~\cite{Xiong_2022_CVPR, ensemble_attack_and_defence_tramr2018ensemble}. 
But unlike they set the surrogate model as the same architecture as the model used in ensemble training,
we set the experiments under a more challenging scenario where we use totally different architectures as surrogate models (\textit{i.e.}, our default settings). As we can see from Table~\ref{tab:comparison_defence}, despite the challenge to attack an adversarially trained black-box model, our AdaEA exhibits the strongest attack performance among the compared methods. For the input transformation-based defenses, we adopt six popular input transformation-based defenses to test the attack performance of each method. From the results in columns seven to thirteen of Table~\ref{tab:comparison_defence}, AdaEA achieves the best results where it surpasses the baseline by $7.9$, $8.27$ and $4.93$ on the base I-FGSM, MI-FGSM, and DI$^2$-FGSM attack, respectively.
\begin{table*}[ht]
\centering
\fontsize{8.5}{10.5}\selectfont
\caption{The robust accuracy (\%) against three adversarial training models and six advanced defense methods on CIFAR-10. The results of input transformation-based defenses are the average results of all target models. The bolded numbers indicate the best results.}
\vspace{5pt}
\label{tab:comparison_defence}
\resizebox{\linewidth}{!}{
\begin{tabular}{ccccccccccccc}
\toprule[1pt]
\multirow{2}{*}{Base} & \multirow{2}{*}{Attack} & \multicolumn{4}{c}{Adversarial training defense} & \multicolumn{7}{c}{Input transformation-based defenses} \\ \cmidrule(lr){3-6} \cmidrule(lr){7-13}
 &  & Inc-v3$_{ens3}$ & Inc-v3$_{ens4}$ & Inc-v2$_{ens}$ & Avg. & R\&P & Bit-R & JPEG & ComDefend & RS & NRP & Avg. \\ 
\toprule[1pt]
\multirow{3}{*}{I-FGSM} & Ens & 0.54 & 0.67 & 0.55 & 0.59 & 18.98 & 32.75 & 23.58 & 83.82 & 57.44 & 13.07 & 38.27 \\
 & SVRE & 0.64 & 0.79 & 0.65 & 0.69 & 20.56 & 35.94 & 26.35 & 83.77 & 57.77 & 12.86 & 39.54 \\
 & AdaEA & \textbf{0.79} & \textbf{0.98} & \textbf{0.79} & \textbf{0.85} & \textbf{26.93} & \textbf{49.67} & \textbf{40.20} & \textbf{84.06} & \textbf{59.65} & \textbf{16.51} & \textbf{46.17} \\ 
 \toprule[1pt]
\multirow{3}{*}{MI-FGSM} & Ens & 0.73 & 0.99 & 0.75 & 0.82 & 26.38 & 43.51 & 36.10 & 83.94 & 58.56 & 5.11 & 42.27 \\
 & SVRE & 0.55 & 0.65 & 0.66 & 0.62 & 16.41 & 25.39 & 23.08 & 83.74 & 56.67 & 3.91 & 34.87 \\
 & AdaEA & \textbf{1.14} & \textbf{1.38} & \textbf{1.21} & \textbf{1.24} & \textbf{37.31} & \textbf{60.90} & \textbf{53.74} & \textbf{84.21} & \textbf{61.64} & \textbf{5.41} & \textbf{50.54} \\ 
 \toprule[1pt]
\multirow{3}{*}{DI$^2$-FGSM} & Ens & 1.47 & 1.72 & 1.79 & 1.66 & 62.92 & 76.80 & 72.54 & 84.16 & 60.96 & 5.30 & 60.44 \\
 & SVRE & 0.85 & 1.02 & 1.01 & 0.96 & 30.77 & 34.46 & 33.79 & 83.77 & 57.75 & 4.28 & 40.80 \\
 & AdaEA & \textbf{2.27} & \textbf{2.49} & \textbf{2.50} & \textbf{2.42} & \textbf{71.83} & \textbf{82.24} & \textbf{79.99} & \textbf{84.37} & \textbf{64.90} & \textbf{8.92} & \textbf{65.37} \\ 
 \bottomrule[1pt]
\end{tabular}
}
\vspace{-10pt}
\end{table*}

\begin{figure}[t]
\begin{center}
\includegraphics[width=0.95\linewidth]{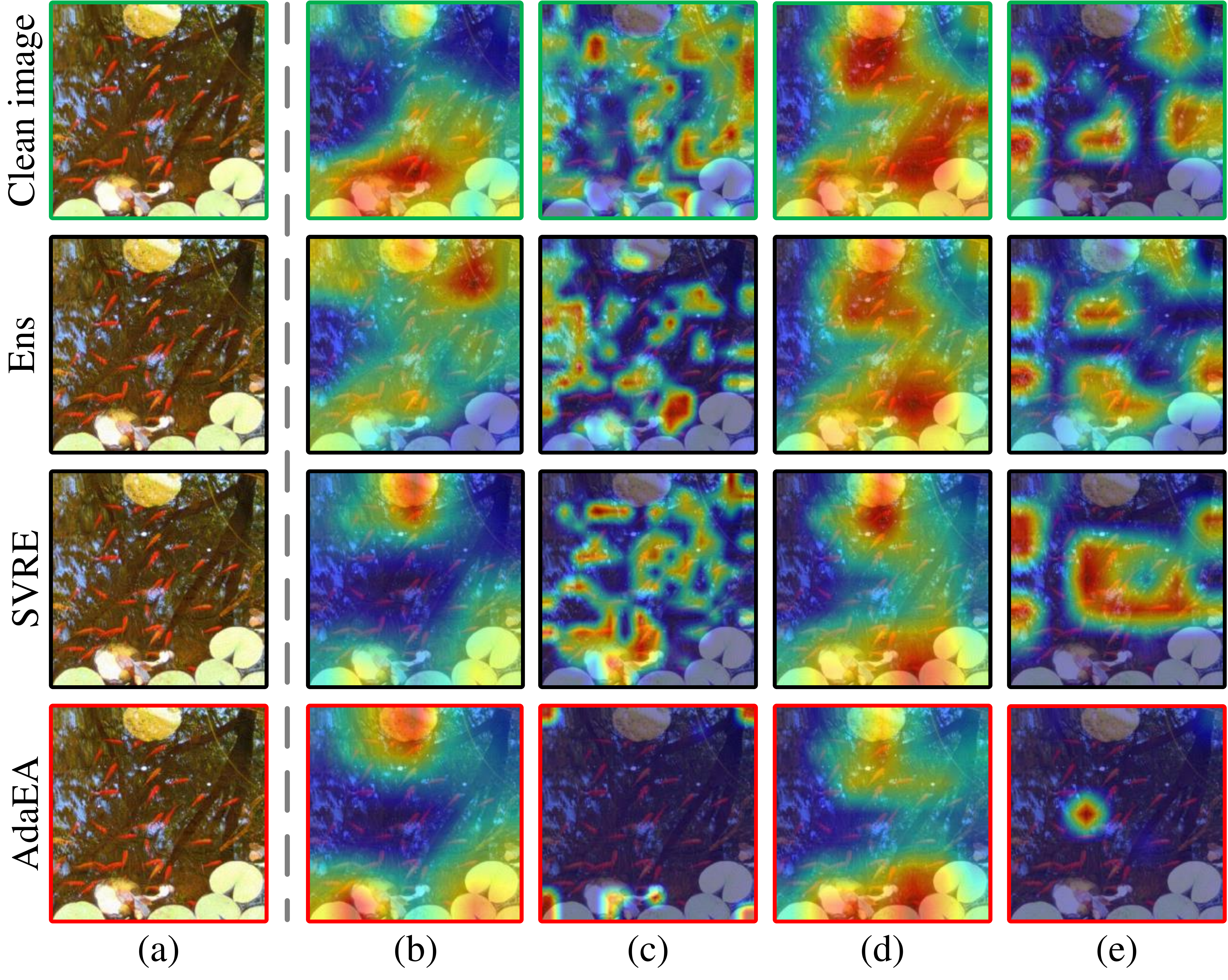}
\end{center}
\vspace{-5pt}
\caption{Heatmaps of different inputs in the surrogate models and black-box models. (a) input images, including clean image and adversarial examples generated by each attack method. (b)-(e) are the heatmaps on the surrogate models (Res-18, ViT-T) and black-box models (WRN50-2, Swin-T), respectively.}
\vspace{-10pt}
\label{fig:heatmaps}
\end{figure}

\noindent\textbf{Visualization of attack performance.} To intuitively show the attack performance, we visualize the heatmaps of clean image and adversarial examples generated by different ensemble methods in both white-box and black-box models in Figure\,\ref{fig:heatmaps}. As can be observed in the Figure\,\ref{fig:heatmaps} (b) and (c), the attention of the white-box models changes on all the generated adversarial images compared with the clean image, which indicates that the generated adversarial examples can effectively trigger the wrong prediction of these models. However, when transferred to black-box models, the Ens and SVRE methods fail to mislead the model attention where the heatmaps are similar to the clean image, as shown in the second to third rows of Figure\,\ref{fig:heatmaps} (d)-(e). In contrast, thanks to the amplification of potential intrinsic adversarial information via AGM-DRF schemes in AdaEA, the generated adversarial example can still fool the attention of black-box models where the attention is dramatically changed in Figure\,\ref{fig:heatmaps} (d)-(e).


\subsection{Ablation Studies}\label{sec:ablation}
In this subsection, we conduct a series of ablation experiments to study the effects of key components and hyper-parameters in our AdaEA.

\vspace{3pt}
\noindent \textbf{On the components of AdaEA.} We first examine the effectiveness of AGM and DRF mechanisms in our AdaEA. Specifically, we perform four ensemble methods: the naive ensemble attack, ensemble with AGM, ensemble with DRF, and our AdaEA involving both AGM and DRF on black-box attacks. The results are reported in Table~\ref{tab:ablation_module}. As can be seen, using AGM can effectively enhance the attack transferability with $12.50\%$ averaging improvements, indicating its effectiveness in amplification of adversarial information during gradient ensemble. It is interesting to see that adding DRF into baseline brings significant improvements on the transferability to ViTs, \textit{i.e.}, $23.94\% \xrightarrow{} 47.57\%$. This is due to the wide differences across CNNs and ViTs, reducing the gradient disparity among the CNNs and ViTs can provide more stable and better attack performance. In general, AGM together with DRF can provide the best transferability with a large improvements over the baseline, \textit{i.e.}, $27.29 \% \xrightarrow{} 44.78 \%$ in average.

\begin{figure}
\begin{center}
\includegraphics[width=1.0\linewidth]{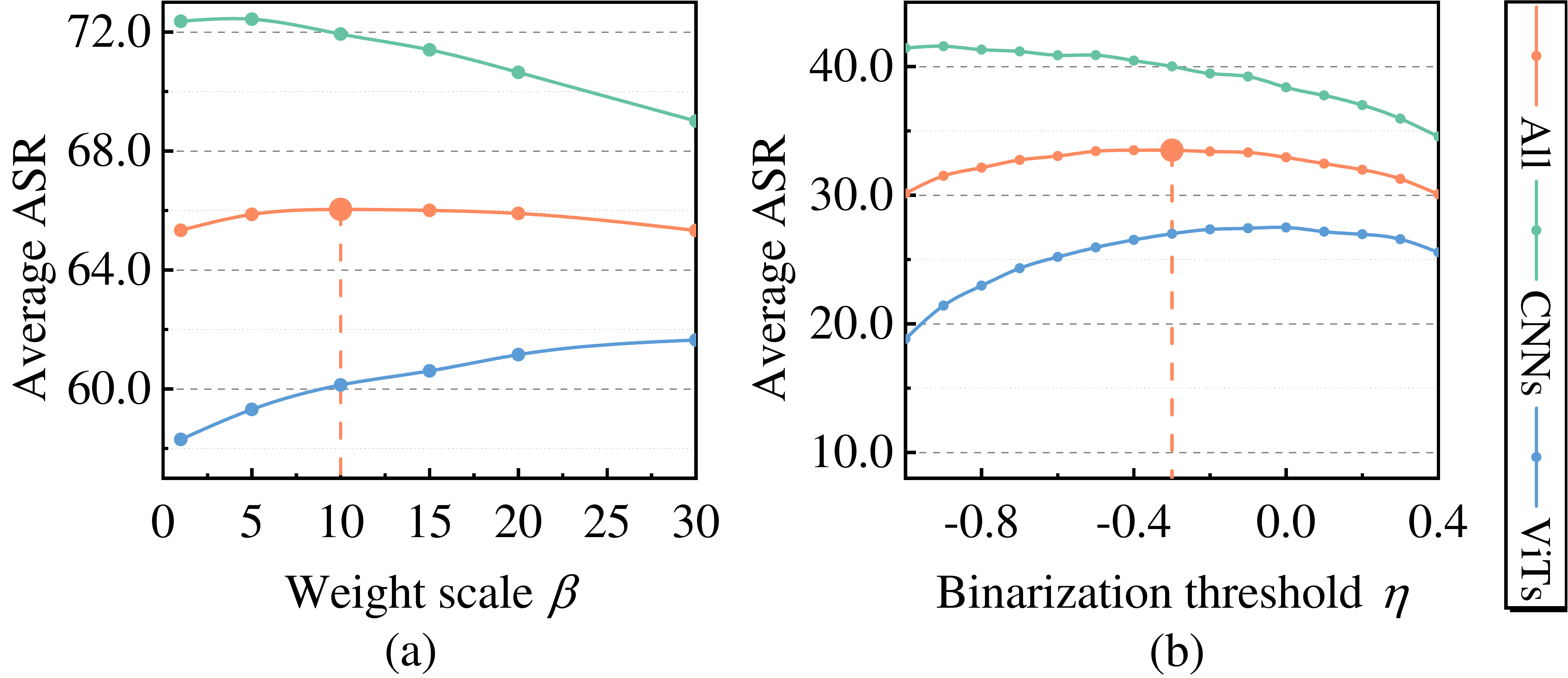}
\end{center}
\vspace{-10pt}
\caption{Ablation study on (a) weighting scale $\beta$ in AGM and (b) binarization threshold $\eta$ in DRF.}
\label{fig:ablation_cos}
\vspace{-15pt}
\end{figure}

\vspace{3pt}
\noindent \textbf{On hyper-parameter sensitivity.} We study the sensitivity of our AdaEA to the weighting scale $\beta$ in Eq.\,\eqref{eq:weight_balance_2} and the binarization threshold $\eta$ in Eq.\,\eqref{eq:drf_3}. We use Res-18 and ViT-T as the surrogate models for ensemble, and show the curves of averaging success rate on black-box CNNs, ViTs, and all the models in Figure\,\ref{fig:ablation_cos}. As we can see in Figure\,\ref{fig:ablation_cos} (a), a larger value of $\beta$ leads to better trasferability to ViTs but lower transferability to CNNs. This suggests that the gradients of ViTs play a critical role in AGM process, a larger $\beta$ can amplify the focus on ViTs. We set $\beta=10$ as the average attack success rate on all the target models reaches the peak at $\beta=10$. For the binarization threshold $\eta$ in Figure\,\ref{fig:ablation_cos} (b), the transferability to ViTs gains large improvements by reducing the disparity as $\eta$ increases, but the transferability to CNNs shows a bit drop. The average performance on all the models increases and reaches the peak at $\eta=-0.3$. 

\begin{table}
\centering
\fontsize{8.5}{10.5}\selectfont
\caption{Experimental results of average attack success rate (\%) on the component ablations in AdaEA.} 
\vspace{2pt}
\label{tab:ablation_module}
\begin{tabular}{ccccc}
\toprule[1pt]
{\begin{tabular}[c]{@{}c@{}}Ens models\end{tabular}} &{Method} & CNNs & ViTs & All ($\Delta$) \\ 
\toprule[1pt]
\multirow{4}{*}{\begin{tabular}[c]{@{}c@{}}Res-18, \\ Inc-v3, \\ ViT-T, \\ DeiT-T\end{tabular}} 
 & Ens & 29.96 & 23.94 & 27.29 \\
 & +AGM & 39.48 & 40.18 & 39.79 (+12.5)  \\
 & +DRF & 38.31 & 47.57 & 42.42 (+15.13) \\ \cmidrule{2-5} 
 & AdaEA & \textbf{40.85} & \textbf{49.69} & \textbf{44.78 (+22.4)} \\ 
 \bottomrule[1pt]
\end{tabular}
\vspace{-1pt}
\end{table}

\begin{table*}
\centering
\fontsize{8.5}{10.5}\selectfont
\caption{Comparison of average attack success rate (\%) between ensemble attack and our AdaEA under different ensemble models on CIFAR-10. Bolded numbers signify better results.}
\label{tab:ablation_model_combination}
\vspace{5pt}
\begin{tabular}{cccccccccccc}
\toprule[1pt]
\multirow{2}{*}{\begin{tabular}[c]{@{}c@{}}Ensemble \\models\end{tabular}} & \multirow{2}{*}{\# CNNs} & \multirow{2}{*}{\# ViTs} & \multirow{2}{*}{Attack} & \multicolumn{4}{c}{CNNs} & \multicolumn{4}{c}{ViTs}\\ \cmidrule(r){5-8} \cmidrule(r){9-12}
 & &  &  & Res-50 & WRN101-2  & BiT-101 & Average & ViT-B & DeiT-B  & Swin-S & Average\\ 
 \toprule[1pt]
 \multirow{2}{*}{\begin{tabular}[c]{@{}c@{}}Inc-v3,\\ DeiT-T\end{tabular}} & \multirow{2}{*}{1} & \multirow{2}{*}{1}
& Ens & 14.52 & 8.09 & {5.73} &9.45 & 3.76 & 7.18  & {8.79} & 6.58 \\
 &  &   & AdaEA & \textbf{29.02} & \textbf{18.53} & {\textbf{19.25}} & \textbf{22.27} & \textbf{20.52} & \textbf{36.75}  & {\textbf{33.09}} & \textbf{30.12} \\ \hline
 
\multirow{2}{*}{\begin{tabular}[c]{@{}c@{}}Res-18,\\ Inc-v3, ViT-T\end{tabular}} & \multirow{2}{*}{2} & \multirow{2}{*}{1} 
& Ens & 43.39 & 22.83  & {13.23} & 26.48 & 5.12 & 10.58 & {21.68} & 12.46\\
 & & & AdaEA & \textbf{49.30} & \textbf{27.03} & {\textbf{16.51}}  & \textbf{30.95}& \textbf{8.47} & \textbf{17.71}  & {\textbf{29.61}} & \textbf{18.60} \\ \hline
 
\multirow{2}{*}{\begin{tabular}[c]{@{}c@{}}Inc-v3,\\ ViT-T, Swin-T\end{tabular}} & \multirow{2}{*}{1} & \multirow{2}{*}{2} 
& Ens & 24.19 & 12.70  & {9.34} & 15.41 & 6.19 & 13.07  & {71.43} & 30.23\\
 &   & & AdaEA & \textbf{39.07} & \textbf{20.86} & {\textbf{16.95}} & \textbf{25.63} & \textbf{15.02} & \textbf{33.33}  & {\textbf{95.66}} & \textbf{48.00} \\ \hline
 
\multirow{2}{*}{\begin{tabular}[c]{@{}c@{}}Res-18, Inc-v3\\ BiT-50\end{tabular}} & \multirow{2}{*}{3}
& \multirow{2}{*}{0} & Ens & 52.86 & 31.69  & {68.21} & 50.92& 4.15 & 7.08  & {21.01} & 10.75\\
 &  &  & AdaEA & \textbf{60.27} & \textbf{37.90}  &{\textbf{72.20}} & \textbf{56.79}  & \textbf{5.28} & \textbf{9.49}  & {\textbf{25.97}} & \textbf{13.58}\\ \hline
 
\multirow{2}{*}{\begin{tabular}[c]{@{}c@{}}ViT-T, DeiT-T,\\ Swin-T\end{tabular}} & \multirow{2}{*}{0} & \multirow{2}{*}{3}
& Ens & \textbf{52.70} & 29.41 & {27.90} & 36.67  & 38.76 & 71.60  & {\textbf{99.00}} & 69.79  \\
 &  &  & AdaEA & 50.14 & \textbf{30.05}  & {\textbf{29.41}} & {\textbf{36.53}} & \textbf{45.92} & \textbf{75.25}  & {97.05} & \textbf{72.74}
\\ \hline
 
\multirow{2}{*}{\begin{tabular}[c]{@{}c@{}}Res-18, Inc-v3,\\ ViT-T, DeiT-T\end{tabular}} & \multirow{2}{*}{2} & \multirow{2}{*}{2} 
& Ens  & 50.42 & 26.85  &{17.61} & 31.63 & 11.59 & 26.15  & {35.42} & 24.39 \\
 &  &  & AdaEA  & \textbf{61.54} & \textbf{38.07}  & {\textbf{28.99}} & \textbf{42.87} & \textbf{31.77} & \textbf{59.72}  & {\textbf{61.38}} & \textbf{50.96} \\ \hline
\multirow{2}{*}{\begin{tabular}[c]{@{}c@{}}Res-18, ViT-T,\\ DeiT-T, Swin-T\end{tabular}} & \multirow{2}{*}{1} & \multirow{2}{*}{3} 
& Ens  & 66.79 & 38.00  & {26.49} & 43.76 & 21.20 & 47.75 & {94.53}  & 54.49 \\
 &  &  & AdaEA & \textbf{71.39} & \textbf{42.88}  & {\textbf{34.70}} & \textbf{49.66}  & \textbf{44.45} & \textbf{76.05} & {\textbf{98.00}} & \textbf{72.83}  \\ \hline
\multirow{2}{*}{\begin{tabular}[c]{@{}c@{}}Res-18, Inc-v3,\\ BiT-50, ViT-T\end{tabular}} & \multirow{2}{*}{3} & \multirow{2}{*}{1} 
& Ens & 61.66 & 37.43  & {72.86} & 57.32 & 9.64 & 18.64  & {39.08}  & 22.45 \\
 &  &  & AdaEA & \textbf{69.91} & \textbf{45.16} & {\textbf{76.39}} & {\textbf{63.82}} & \textbf{14.64} & \textbf{27.88}  & {\textbf{49.15}} & \textbf{30.56}\\ 
 \bottomrule[1pt]
\end{tabular}
\vspace{-5pt}
\end{table*}

\subsection{Further Analysis}\label{sec:analysis}
Since our work is among the first grups to study the adversarial transfer across both CNNs and ViTs, we further analyze the transferability of adversarial examples from the perspective of surrogate models used during the ensemble by considering the following questions. 

\vspace{3pt}
\noindent \textbf{What effect does the number of surrogate models have on the transferability?} 
We first test the effect of different numbers of surrogate models on the ensemble attack performance. 
From Table~\ref{tab:ablation_model_combination}, we can see that as the number of surrogate model increases, the overall attack success rate improves from the first row to the bottom row. The ensemble using four surrogate models improves the average success rate by around $20\%$ on both CNNs and ViTs over using two surrogate models, as can been seen in the second and seventh rows of Table~\ref{tab:ablation_model_combination}. Intuitively, using more surrogate models can lead to better transferability since more adversarial information can be captured. More importantly, our AdaEA consistently improves the ensemble attack performance regardless the number of ensemble models.

\vspace{3pt}
\noindent \textbf{How does different proportions of CNNs to ViTs in surrogate models affect the overall transferability?} As CNNs and ViTs are two main branches in the family of DNNs, we investigate the effects of the proportions of CNNs to ViTs in the surrogate models on the overall transferability. By observing the second, third, and ninth rows of Table\,\ref{tab:ablation_model_combination}, as the number of CNNs increases in the surrogate models, the attack rate on CNNs obviously improves. But in contrast, the attack success rate on ViTs is not going higher. This indicates that the ensemble gradient focuses more on the gradients of CNNs when the CNNs dominate in the surrogate models.
When only CNNs are used as surrogate models in the fifth row of Table\,\ref{tab:ablation_model_combination}, the attack has high success rates on CNNs but reveals a low transfer rate on ViTs. 
But interestingly, when the proportion of CNNs to ViTs becomes $0:3$ in the sixth row of Table\,\ref{tab:ablation_model_combination}, where only ViTs are used, the ensemble attack still exhibits a high transfer rate to CNNs. The same results can be seen in the fourth and eighth rows of Table\,\ref{tab:ablation_model_combination} when the ViTs dominate the surrogate models, the transfer to CNNs can still maintain a high attack success rate. This phenomenon indicates that \textit{it is easier to transfer attacks from ViTs to CNNs compared with transferring from CNNs to ViTs}. We attribute this to the more complex architecture and global modeling ability of ViTs, which makes ViTs capable of extracting more generic adversarial information.

\section{Conclusion}
In this work we propose AdaEA, an adaptive ensemble adversarial attack that merges the gradients of surrogate models via monitoring on the contribution of each model to the overall adversarial objective, for boosting the transferability of adversarial examples. We show that AdaEA can effectively enhance the adversarial transferability across models with a large margin over the existing ensemble methods under various settings, even those with wide architecture differences, \textit{e.g.}, CNNs and ViTs, which demonstrates the effectiveness of our method in capturing intrinsic adversarial information of inputs.

\section{Acknowledgements}
This work was partly supported by the National Natural Science Foundation of China under Grant Nos. 62202104, 62102422, 62072109 and U1804263; the Ministry of Science and Technology, Taiwan, under Grant MOST 111-2628-E-155-003-MY3; and Youth Foundation of Fujian Province, P.R.China, under Grant No.2021J05129.

{\small
\bibliographystyle{ieee_fullname}
\bibliography{egbib}
}

\end{document}